\documentclass[
  journal=proceedings,
  manuscript=article-type,
  year=2025
]{PMET_proc}

\usepackage{amsmath}
\usepackage[nopatch]{microtype}
\usepackage{booktabs}
\usepackage{orcidlink}

\newcommand{\orcidauthor}[2]{#1~\orcidlink{#2}}

\title{Optimizing the Landscape of LLM Embeddings with Dynamic Exploratory Graph Analysis for Generative Psychometrics: A Monte Carlo Study}

\author{\orcidauthor{Hudson Golino}{0000-0002-1601-1447}}
\affiliation{University of Virginia, Charlottesville, VA, USA}
\email{hfg9s@virginia.edu}


\begin{document}

\begin{abstract}
Large language model (LLM) embeddings are increasingly used to estimate dimensional structure in psychological item pools prior to data collection, yet current applications treat embeddings as static, cross-sectional representations. This approach implicitly assumes uniform contribution across all embedding coordinates and overlooks the possibility that optimal structural information may be concentrated in specific regions of the embedding space. This study reframes embeddings as searchable landscapes and adapts Dynamic Exploratory Graph Analysis (DynEGA) to systematically traverse embedding coordinates, treating the dimension index as a pseudo-temporal ordering analogous to intensive longitudinal trajectories.

A large-scale Monte Carlo simulation embedded items representing five dimensions of grandiose narcissism using OpenAI's \texttt{text-embedding-3-small} model, generating millions of network estimations across systematically varied item pool sizes (3-40 items per dimension) and embedding depths (3-1,298 dimensions). Results reveal that Total Entropy Fit Index (TEFI) and Normalized Mutual Information (NMI) leads to competing optimization trajectories across the embedding landscape. TEFI achieves minima at deep embedding ranges (900--1,200 dimensions) where entropy-based organization is maximal but structural accuracy degrades, whereas NMI peaks at shallow depths where dimensional recovery is strongest but entropy-based fit remains suboptimal. Single-metric optimization produces structurally incoherent solutions, whereas a weighted composite criterion (70\% NMI, 30\% TEFI) identifies embedding dimensions depth regions that jointly balance accuracy and organization.

Landscape optimization via DynEGA consistently outperforms standard cross-sectional Exploratory Graph Analysis across all item pool sizes, with greatest gains observed for pools of more than 15 items per dimension. Optimal embedding depth scales systematically with item pool size. These findings establish embedding landscapes as non-uniform semantic spaces requiring principled optimization rather than default full-vector usage, positioning DynEGA as a foundational tool for embedding-based dimensionality assessment in generative psychometrics.
\end{abstract}

\section{Introduction}

LLM-based embeddings have become central to text analytics, including semantic search, clustering, topic modeling, and automated scale development. In psychological measurement, generative pipelines such as AI-GENIE use LLMs to generate candidate items and rely on embeddings to estimate dimensionality and structural organization prior to data collection \citep{russell_2024aigenie}.

Text embeddings map language units into dense vectors in high-dimensional space, such that semantically similar texts occupy nearby positions. This geometric representation allows researchers to compute similarity relations between items and apply psychometric tools directly to semantic content rather than response data alone. Recent work has increasingly treated embeddings as pre-empirical representations of item meaning, enabling dimensionality assessment, content validation, and scale refinement before administration \citep{russell_2024aigenie, wulff2025embeddinga}.

Simulation evidence further suggests that methodological choices at this stage are consequential. \citet{garrido2025estimating} demonstrated that principal component analysis applied to item embeddings systematically overestimates dimensionality, whereas network-based approaches recover the generating structure with substantially greater accuracy. These findings motivate a shift away from treating embeddings as static vectors to be analyzed wholesale, toward approaches that explicitly model structure, heterogeneity, and organization within the embedding space.

The present study extends this emerging literature by conceptualizing embedding spaces as searchable landscapes. Rather than assuming that all embedding coordinates contribute uniformly to dimensional structure, the proposed approach uses Dynamic Exploratory Graph Analysis to traverse the embedding dimension index and identify regions that optimally balance structural accuracy and entropy-based fit.

\section{Background}

\subsection{Exploratory Graph Analysis and Dynamic Extensions}

Exploratory Graph Analysis (EGA) estimates dimensional structure by modeling item associations as a network and identifying communities using graph-theoretic algorithms \citep{golino2017ega1, golino2019investigating, christensen2020comparing}. In EGA, variables are represented as nodes in a network, and their partial correlations (after controlling for all other variables) are represented as edges. Community detection algorithms then identify clusters of variables that are more strongly connected to each other than to variables in other clusters.

Dynamic EGA (DynEGA) generalizes this framework to ordered, (intensive) longitudinal time-series data by combining state-space reconstruction, derivative estimation via Generalized Local Linear Approximation \citep[GLLA, see:][]{glla2010}, and exploratory graph analysis \citep{golino2020modeling, tomasevic2024decoding}. The method operates in three stages. First, each time series is transformed into a time-delay embedding matrix through state-space reconstruction \citep{rosenstein1993practical}. For a univariate time series $U = \{x_t, x_{t+1}, \ldots, x_{t+N}\}$, the time-delay embedding matrix is constructed as:
\begin{equation}
\mathbf{X} = \begin{bmatrix}
x_t & x_{t+\tau} & \cdots & x_{t+(n-1)\tau} \\
x_{t+1} & x_{t+1+\tau} & \cdots & x_{t+1+(n-1)\tau} \\
\vdots & \vdots & \ddots & \vdots \\
x_{t+M-1} & x_{t+M-1+\tau} & \cdots & x_{t+M-1+(n-1)\tau}
\end{bmatrix},
\end{equation}
where $\tau$ is the reconstruction delay (lag), $n$ is the embedding dimension, and $M = N - (n-1)\tau$ is the number of embedded observations.

Second, derivatives are estimated using GLLA. The derivative matrix $\mathbf{Y}$ is computed as:
\begin{equation}
\mathbf{Y} = \mathbf{X}\mathbf{L}(\mathbf{L}^{\prime}\mathbf{L})^{-1},
\end{equation}
where $\mathbf{L}$ is a weight matrix. Each column of $\mathbf{L}$ corresponds to a derivative order $\alpha$ (where $\alpha = 0$ represents the observed values, $\alpha = 1$ the first derivative or velocity, and $\alpha = 2$ the second derivative or acceleration):
\begin{equation}
\mathbf{L}_{\alpha} = \frac{[\Delta_t(v - \bar{v})]^{\alpha}}{\alpha!},
\end{equation}
where $\Delta_t$ is the time between observations, $v = [1, 2, \ldots, n]$ indexes the embedded dimensions, and $\bar{v}$ is the mean of $v$. This procedure transforms each time series into a matrix of derivatives capturing how variables change over time.

Third, after computing derivatives for all variables, the resulting derivative matrices are column-bound to form a single matrix for network estimation. EGA is then applied to this derivative space to identify clusters of variables that are changing together \citep{golino2020modeling}. Unlike standard EGA, which operates on raw covariances, DynEGA operates on derivatives, meaning that edges in the network represent coordinated patterns of change rather than static associations.

In the DynEGA framework, an important hyperparameter is the number of embedding dimensions $n$, which determines the size of the temporal window used to compute derivatives. This parameter can be optimized using fit indices such as the Total Entropy Fit Index \citep[TEFI;][]{golino2019entropy}.

\subsection{Total Entropy Fit Index}

The Total Entropy Fit Index (TEFI) quantifies model fit using concepts from quantum information theory, capturing the degree of structural organization present in multivariate systems \citep{golino2019entropy}. TEFI is based on Von Neumann entropy \citep{vonneumann1927}, a measure of disorder or uncertainty in a system adapted from quantum mechanics to correlation matrices.

To compute TEFI, the correlation matrix $\mathbf{R}$ is first transformed into a density-like matrix $\boldsymbol{\rho}$ by normalizing such that the trace equals one:
\begin{equation}
\boldsymbol{\rho} = \frac{\mathbf{R}}{\text{tr}(\mathbf{R})}.
\end{equation}
This transformation allows the correlation matrix to satisfy the mathematical properties of a quantum density matrix. Von Neumann entropy is then computed as:
\begin{equation}
\mathcal{S}(\boldsymbol{\rho}) = -\text{tr}(\boldsymbol{\rho} \log \boldsymbol{\rho}),
\end{equation}
which can be equivalently expressed using the eigenvalues  $\lambda_1, \lambda_2, \ldots, \lambda_m$ of $\boldsymbol{\rho}$ \citep{wihler2014computing}:
\begin{equation}
\mathcal{S}(\boldsymbol{\rho}) = -\sum_{i=1}^{m} \lambda_i \log \lambda_i.
\end{equation}

Von Neumann entropy \citep{vonneumann1927} is an index that was developed to quantify the amount of disorder in a system. It's also been used to quantify the entanglement between two subsystems in quantum physics \citep{quantuminfo}, which occurs when two (or more) particles become inextricably linked. Lower entropy indicates greater structural coherence and organization.

The TEFI evaluates a proposed dimensional partition by comparing the average entropy of individual dimensions to the total system entropy.  Given a partition with $N_F$ dimensions, TEFI is computed as:
\begin{equation}
\text{TEFI} = \left[\frac{\sum_{k=1}^{N_F} \mathcal{S}(\boldsymbol{\rho}_k)}{N_F} - \mathcal{S}(\boldsymbol{\rho})\right] + \left[\left(\mathcal{S}(\boldsymbol{\rho}) - \sum_{k=1}^{N_F} \mathcal{S}(\boldsymbol{\rho}_k)\right) \times \sqrt{N_F}\right],
\end{equation}

where $\mathcal{S}(\boldsymbol{\rho}_k)$ is the Von Neumann entropy of the density submatrix corresponding to dimension $k$, and $\mathcal{S}(\boldsymbol{\rho})$ is the total entropy of the full system.

The first bracketed term represents the difference between the average entropy of the dimensions and the total entropy. This term decreases as dimensions become more coherent and internally organized. The second bracketed term penalizes solutions with excessive dimensionality, weighted by $\sqrt{N_F}$ to control for the growth trajectory as the number of dimensions increases. Together, these components balance within-dimension coherence against model complexity \citep{golino2019entropy}.

Lower (more negative) TEFI values indicate better fit, reflecting dimensional solutions that reduce uncertainty while avoiding overfitting. TEFI operates under a simple structure assumption, where each variable belongs to exactly one dimension. Misallocation of variables increases entropy and thus increases TEFI, making the index sensitive not only to the number of dimensions but also to their composition \citep{golino2019entropy, golino2024gentefi}.

Together, DynEGA and TEFI provide complementary information about structural accuracy and system organization. DynEGA identifies which variables change together in a dynamic system with underlying dynamic dimensions, while TEFI quantifies how well a proposed partition organizes the correlation structure.

\subsection{Embedding Landscapes as Pseudo-Dynamic Systems}

In traditional DynEGA applications, time defines the ordering that generates trajectories. For LLM embeddings, the embedding dimension index can play an analogous role. Each item embedding can be conceptualized as a trajectory through a high-dimensional semantic landscape, where the coordinate index acts as pseudo-time. Traversing this index allows researchers to identify regions of the embedding space where dimensional structure is most coherent.

This perspective aligns embedding-based psychometrics with broader developments in network modeling and dynamic systems analysis. Rather than treating all 1,536 dimensions of an OpenAI embedding as equally informative via a cross-sectional analysis \citep[common approach nowadays in Generative Psychometrics, see:][]{ russell_2024aigenie, garrido2025estimating}, DynEGA can systematically search across increasing depths to locate subspaces that optimize both structural recovery (measured by normalized mutual information) and entropy-based fit (measured by TEFI).

\section{Methods}

\subsection{Conceptual Domain and Item Generation}

The simulations focused on grandiose narcissism, comprising five theoretically motivated dimensions: Authority, Exhibitionism, Superiority, Entitlement, and Exploitativeness \citep{pincus2010pathological}. A pool of 200 candidate items (40 per dimension) was generated using the AI-GENIE framework \citep{russell_2024aigenie}, ensuring controlled semantic variation within and across dimensions. Items were designed to reflect prototypical expressions of each dimension while maintaining sufficient diversity to test the embedding models' capacity to recover dimensional boundaries.

\subsection{Embedding Model and Simulation Design}

Items were embedded using OpenAI’s \texttt{text-embedding-small} model \citep{OpenAIembeddings}, which produces 1,536-dimensional representations. The Monte Carlo simulation systematically varied the number of items per dimension ($k = 3, 4, 5, \ldots, 40$). For each condition, Dynamic Exploratory Graph Analysis (DynEGA) was applied across progressively increasing embedding depths, ranging from 3 to 1,298 dimensions in increments of five. At each embedding depth, Normalized Mutual Information (NMI) and the Total Entropy Fit Index (TEFI) were computed to quantify structural recovery and entropy-based organization, respectively.

Normalized mutual information (NMI) quantifies the correspondence between the estimated dimensional structure and the known true structure, ranging from 0 (no correspondence) to 1 (perfect recovery). TEFI quantifies the entropy-based organization of the estimated structure, with lower values indicating better fit. The simulation design enabled systematic mapping of how these two metrics evolve as a function of both item pool richness and embedding depth.

For each combination of item count and embedding depth, network estimation was performed using the Triangulated Maximally Filtered Graph \citep[TMFG:][]{massara2016network} method within EGA. Community detection was applied using the Walktrap algorithm \citep{pons2006walktrap}, and the resulting partition was evaluated using both NMI and TEFI.

To identify optimal regions in the embedding landscape, a composite metric was constructed as a weighted combination of NMI and TEFI. After normalizing both metrics to a common scale, the composite was defined as $C = 0.70 \times \text{NMI} - 0.30 \times \text{TEFI}_{\text{norm}}$, where the negative sign for TEFI reflects that lower TEFI values indicate better fit. This weighting reflects the prioritization of structural accuracy while maintaining entropy-based organization. 

The reason to use a composite metric is because NMI and TEFI are two metrics with distinct characteristics, leading to different and often competing optima across the embedding landscape. Specifically, NMI is optimized by configurations that maximize agreement with the known dimensional structure, whereas TEFI is optimized by configurations that minimize informational disorder in the estimated network. As a result, optimizing either metric in isolation can yield solutions that are highly accurate but poorly organized, or well-organized but structurally inaccurate. The composite metric resolves this trade-off by identifying regions of the embedding space that jointly balance accuracy and organization, providing a principled criterion for selecting embedding depths that are both interpretable and psychometrically coherent. This is a quick-solution to a typical multi-objective optimization. 

Finally, structural accuracy obtained from the original cross-sectional use of embeddings via Exploratory Graph Analysis \citep[EGA][]{russell_2024aigenie, garrido2025estimating} was compared with the accuracy achieved through joint NMI–TEFI optimization using Dynamic Exploratory Graph Analysis (DynEGA) to search the embedding landscape.

\section{Data and Code Availability}

All R code used to generate the simulations and analyses reported in this paper is publicly available on the Open Science Framework (OSF) at
\url{https://osf.io/4y3vk/overview?view_only=d1d10b0e6d804ee18252cf668521df49}.

\section{Results}

\subsection{Single Example of a Landscape Search}

\begin{figure}[!ht]
  \centering
  \includegraphics[width=\linewidth]{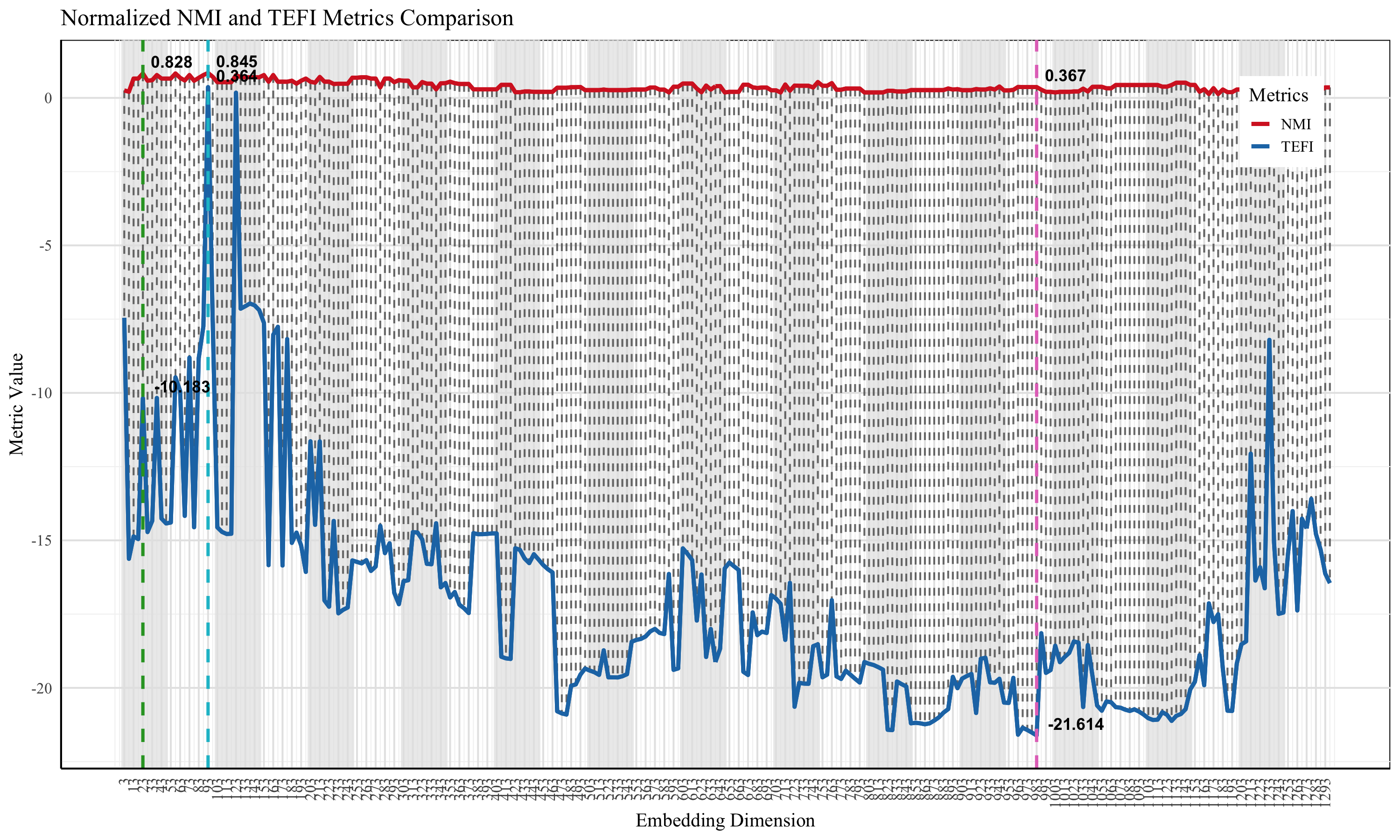}
\caption{Example of the embedding landscape search illustrating NMI (red line) and TEFI (blue line) trajectories across embedding dimensions and the resulting composite optimum (dashed vertical green line).}
  \label{fig:single-landscape-search}
\end{figure}

Figure ~\ref{fig:single-landscape-search} illustrates a single example of the embedding landscape search procedure for the condition with four items per dimension. For each embedding depth, both Normalized Mutual Information (NMI; red line) and the Total Entropy Fit Index (TEFI; blue line) were computed. NMI captures agreement between the recovered and generating dimensional structures, whereas TEFI quantifies entropy-based network organization, with lower values indicating better fit. The two metrics exhibit distinct and partially competing trajectories across the embedding dimension index: NMI is maximized at a shallow depth (93 dimensions; NMI = 0.845), whereas TEFI is minimized at a much deeper region of the embedding space (983 dimensions; TEFI = -21.614).

If NMI were optimized in isolation (dashed vertical blue line), the corresponding TEFI value at that solution would be 0.364, indicating poor entropy-based organization. Conversely, if TEFI were optimized in isolation (dashed vertical pink line), the corresponding NMI value would be 0.367, reflecting weak structural recovery. To resolve this trade-off, a weighted composite score prioritizing structural accuracy (70\% NMI, 30\% TEFI) was used. This composite criterion identifies an optimal solution at 23 embedding dimensions (dashed vertical green line), corresponding to a region of the embedding landscape that balances an acceptable structural recovery (NMI = 0.828) with a good entropy-based organization (TEFI = -10.183).

This example illustrates how embedding landscapes contain multiple local optima and motivates the use of a composite criterion for principled selection of embedding depth in the presence of competing objectives.

\subsection{Vector Field of Total Entropy Fit Index and Normalized Mutual Information} 

\begin{figure}[!ht]
  \centering
  \includegraphics[width=\linewidth]{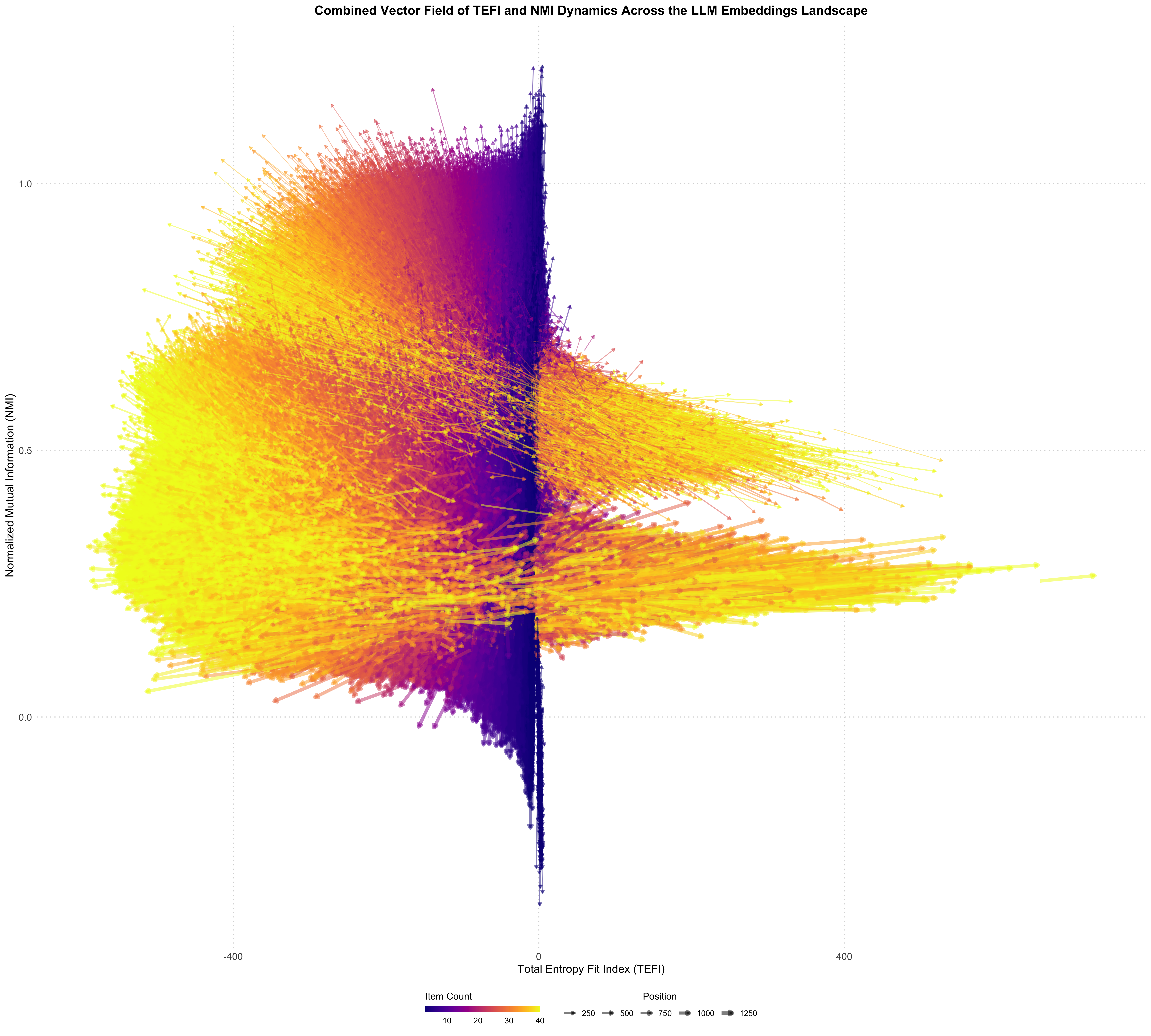}
\caption{Vector field of TEFI-NMI dynamics across the embedding landscape, with arrows indicating local GLLA-based first-order derivatives, color denoting item count (3 items = dark blue, 40 items = yellow), and thickness reflecting embedding position.}
  \label{fig:vector_field_plot}
\end{figure}

Figure ~\ref{fig:vector_field_plot} presents a combined vector field representation of the joint dynamics of the Total Entropy Fit Index (TEFI; x-axis) and Normalized Mutual Information (NMI; y-axis) across the LLM embedding landscape, aggregated over 500 Monte Carlo iterations and item counts ranging from 3 to 40. Each arrow represents a local first-order derivative estimate obtained via Generalized Local Linear Approximation, capturing the instantaneous direction and magnitude of change in TEFI and NMI as embedding depth increases. Arrow color encodes the number of items per dimension (3 items = dark blue, 40 items = yellow), whereas arrow thickness reflects position along the embedding landscape, with thicker arrows corresponding to later embedding coordinates.

Three dynamical regimes can be seen in Figure ~\ref{fig:vector_field_plot} (1) an optimal zone in the upper-left quadrant characterized by thin arrows, moderate item counts, and joint optimization of both metrics; (2) a central transitional region where sparse item pools converge at moderate performance levels; and (3) extreme regions where thick arrows indicate instability arising from either shallow embeddings (positive TEFI) or deep embeddings with higher item counts (extreme negative TEFI with degraded NMI).

The divergent fan pattern in the negative TEFI region demonstrates why single-metric optimization is problematic. Trajectories that follow TEFI improvement alone move toward deep embeddings where structural accuracy degrades, whereas trajectories optimizing NMI alone may settle in regions with suboptimal entropy-based fit. The concentration of thin arrows in the optimal zone points that the composite metric successfully deals with this trade-off by identifying intermediate-depth solutions where both objectives are balanced.

\subsection{Normalized Mutual Information Across Embedding Dimensions} 

\begin{figure}[!ht]
  \centering
    \includegraphics[width=\linewidth]{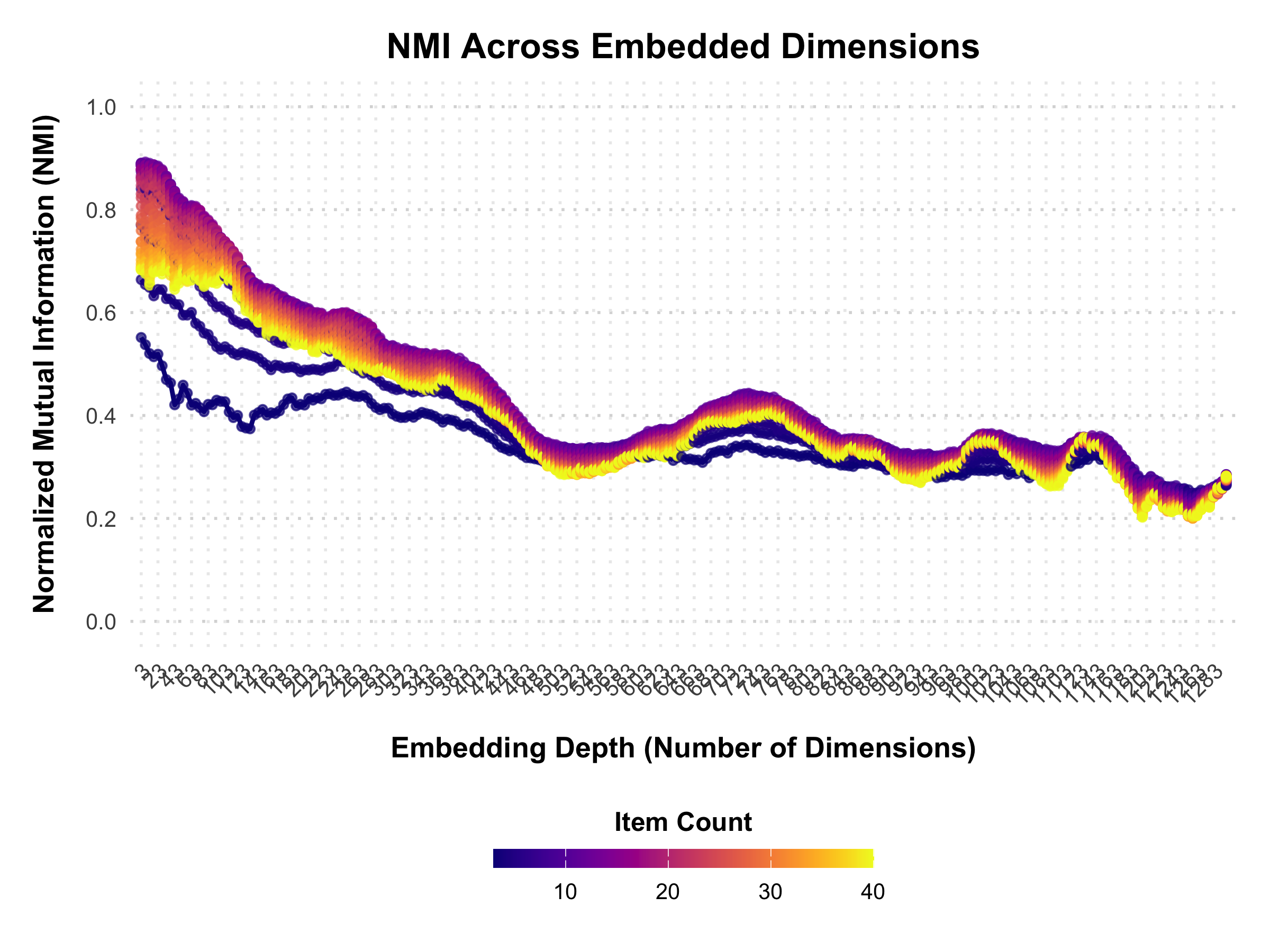}
\caption{Normalized Mutual Information (NMI; y-axis) across embedding depth by number of items per dimension (color), and number of embedding dimensions used (x-axis). NMI measures correspondence between estimated and generating dimensional structures (0 = no correspondence, 1 = perfect recovery). Color gradient indicates item pool size (dark blue = 3 items per dimension, yellow = 40 items per dimension)}
  \label{fig:nmi_plot}
\end{figure}

Figure ~\ref{fig:nmi_plot} shows the distribution of Normalized Mutual Information (NMI) across the embedding landscape for the OpenAI \texttt{text-embedding-small} model, stratified by item count. At shallow embedding depths (approximately 3–50 dimensions), NMI is moderately high to high across item counts, except for the three items per dimension condition. That indicates that early embedding coordinates capture semantic structure sufficient for recovering the generating dimensional organization. As embedding depth increases, NMI declines nonlinearly, revealing a systematic degradation in structural recovery as progressively higher number of embedding dimensions are used.

A broad plateau-like area emerges at intermediate embedding depths (approximately 423-623 dimensions), where NMI stabilizes across item counts, but NMI is very low. This region suggests a regime in which additional embedding dimensions contribute limited gains or losses in structural accuracy. Beyond this plateau, NMI increases, and then declines, with sharper drops observed at larger embedding depths, consistent with the introduction of semantic noise or increasingly fine-grained distinctions that do not align with the dimensional structure of the construct.

Item count modulates this trajectory in a systematic way. Fewer items per dimension exhibit greater variability and lower overall NMI across the embedding landscape, whereas more items per dimension (between 10 and 20) achieve consistently higher NMI values, particularly at shallow and intermediate depths. Too many items per dimensions (e.g., 40), however, degrades NMI irrespective of the number of embedded dimensions used, but with higher NMI at shallow depths. As embedding depth increases, however, NMI basically converges across item counts, suggesting that excessive number of embedding dimensions decreases the accuracy obtained with higher number of items per dimension.

Overall, the figure demonstrates that structural recovery as measured by NMI is strongest at relatively shallow to embedding depths (3-50) and deteriorates as additional embedding dimensions are incorporated. It also shows that using between 10 and 20 items per dimension leads to higher NMI values at shallow embedding depth. These patterns motivate treating embedding depth as a tunable parameter rather than a fixed property and highlight the importance of balancing representational richness against structural fidelity when using LLM embeddings for dimensionality assessment.

\subsection{Total Entropy Fit Index Across Embedding Dimensions}

\begin{figure}[!ht]
  \centering
    \includegraphics[width=\linewidth]{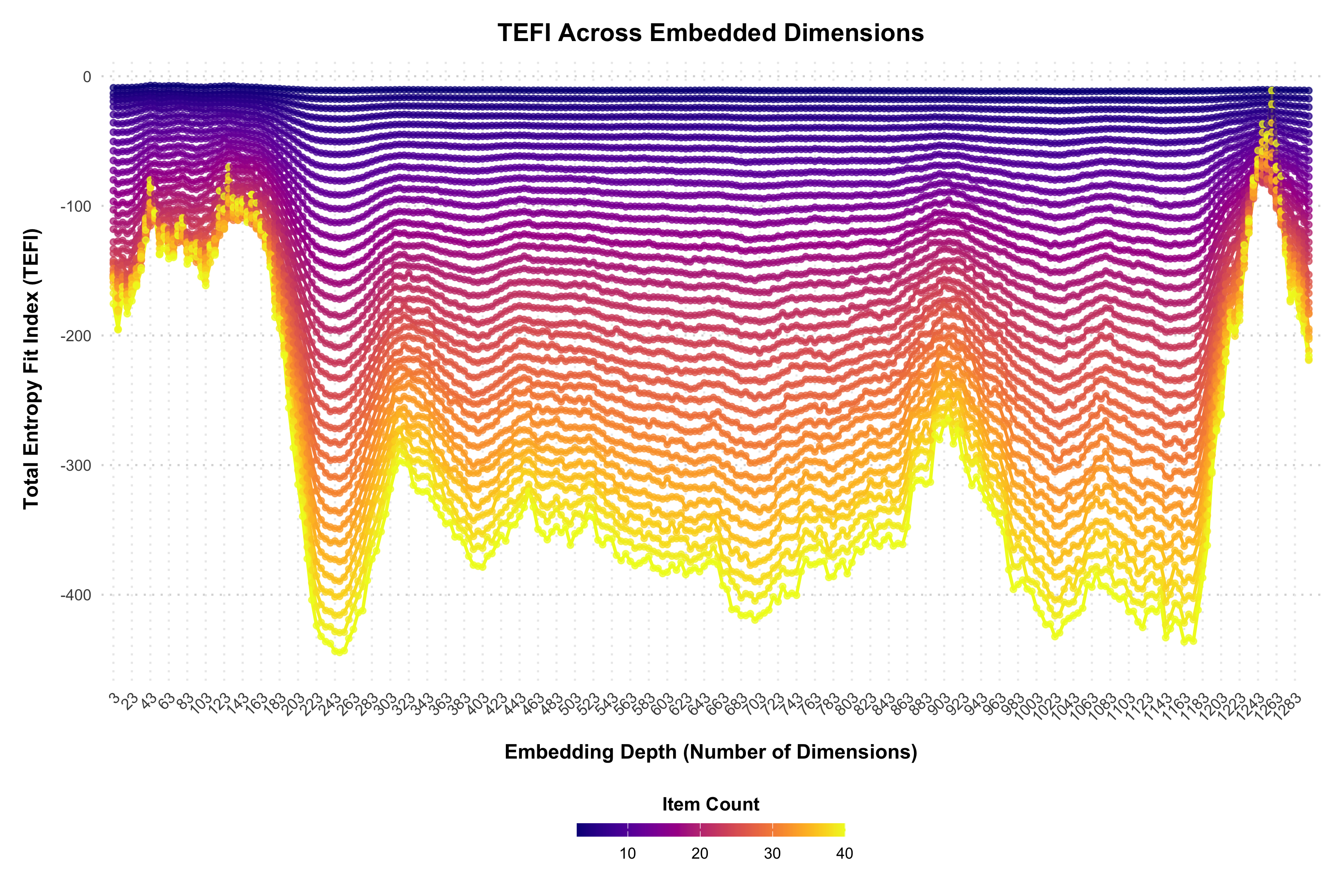}
\caption{Total Entropy Fit Index (TEFI) across embedding depth by item pool size. Lower (more negative) TEFI values indicate better entropy-based fit and greater structural organization. Color gradient indicates item pool size (dark blue = 3 items per dimension, yellow = 40 items per dimension).}
  \label{fig:tefi_plot}
\end{figure}

Figure~\ref{fig:tefi_plot} displays the Total Entropy Fit Index (TEFI) across the embedding landscape for the OpenAI \texttt{text-embedding-small} model, by item count.  Lower (more negative) TEFI values indicate better entropy-based fit, reflecting greater structural organization and reduced uncertainty in the dimensional partition. The TEFI trajectories reveal a strikingly different pattern than NMI, demonstrating that entropy-based organization and structural accuracy follow distinct optimization paths across the embedding landscape.

At very shallow embedding depths (3-183 dimensions), TEFI values are not large negative, across all item counts. As embedding depth increases, TEFI improves (becomes more negative) systematically. The rate and magnitude of improvement depend critically on the number of items per dimension.

The most striking feature of the TEFI landscape is its periodic oscillation pattern, characterized by multiple local minima (optimal regions) and maxima (suboptimal regions) that repeat across the embedding depth range. The first major trough appears at approximately 183-303 dimensions, where TEFI achieves initial local optima, particularly for item pools with 25-40 items per dimension (yellow/orange trajectories). This is followed by a relative maximum (degradation) around 323-643 dimensions, then a second, deeper trough at approximately 663--743. Additional oscillations continue throughout deeper ranges, with notable deep troughs around 963-1063 dimensions and near the maximum depth at 1163 dimensions.

The number of items per dimension has a substantial and systematic effect on TEFI trajectories. Fewer items per dimension (3-8 items; dark purple/purple trajectories) remain closer to zero throughout the landscape, never achieving strong negative TEFI values regardless of embedding depth. This pattern reflects insufficient item coverage to establish coherent within-dimension structure. Item pools with 10-20 items per dimension (pink/salmon trajectories) show more stronger negative TEFI values between -100 and -250 at optimal depths. However, item pools with 25-40 items per dimension (orange/yellow trajectories) demonstrate dramatically lower TEFI values below -300 and reaching as low as -450 in optimal depth ranges.

The vertical stratification of trajectories by number of items per dimension reveals that item pool richness fundamentally determines the achievable level of entropy-based organization. Whereas NMI showed an inverted-U relationship with item count (with 10-20 items optimal), TEFI exhibits a monotonic improvement with increasing item count throughout most of the landscape. This contrast highlights the distinct measurement properties of the two indices. NMI prioritizes accurate dimensional recovery and can be disrupted by using too many items per dimension, whereas TEFI rewards within-dimension coherence and benefits from semantic density.

The TEFI trajectory also differs from the NMI trajectory in its behavior at deep embedding ranges. Where NMI consistently degrades beyond 500 embedding dimensions, TEFI continues to improve or maintains strong performance throughout deeper ranges, particularly for large number of items per dimension. The final spike toward zero at the extreme right (around dimension 1223-1263) represents boundary effects at the maximum dimensionality of the embedding model. This divergence between TEFI and NMI at deep embeddings directly demonstrates the competing optimization objectives identified in the vector field analysis. Following TEFI improvement alone leads to deep embeddings with excellent entropy-based fit but degraded structural accuracy (low NMI), whereas following NMI alone leads to shallow depths with good structural recovery but suboptimal entropy-based organization.

The periodic oscillation pattern observed in TEFI trajectories likely reflects architectural properties of the transformer model underlying the embeddings. Transformer models organize semantic information across layers and attention heads in complex, non-uniform ways, and the oscillations may correspond to depth ranges where the representational structure aligns more or less coherently with the dimensional organization of the psychological construct. Alternatively, the periodicity may reflect interactions between the number of dimensions in the construct (five in this simulation) and the way semantic information is distributed across embedding coordinates.

\subsection{Optimization using Total Entropy Fit Index and Normalized Information Index}

\begin{figure}[!ht]
  \centering
    \includegraphics[width=\linewidth]{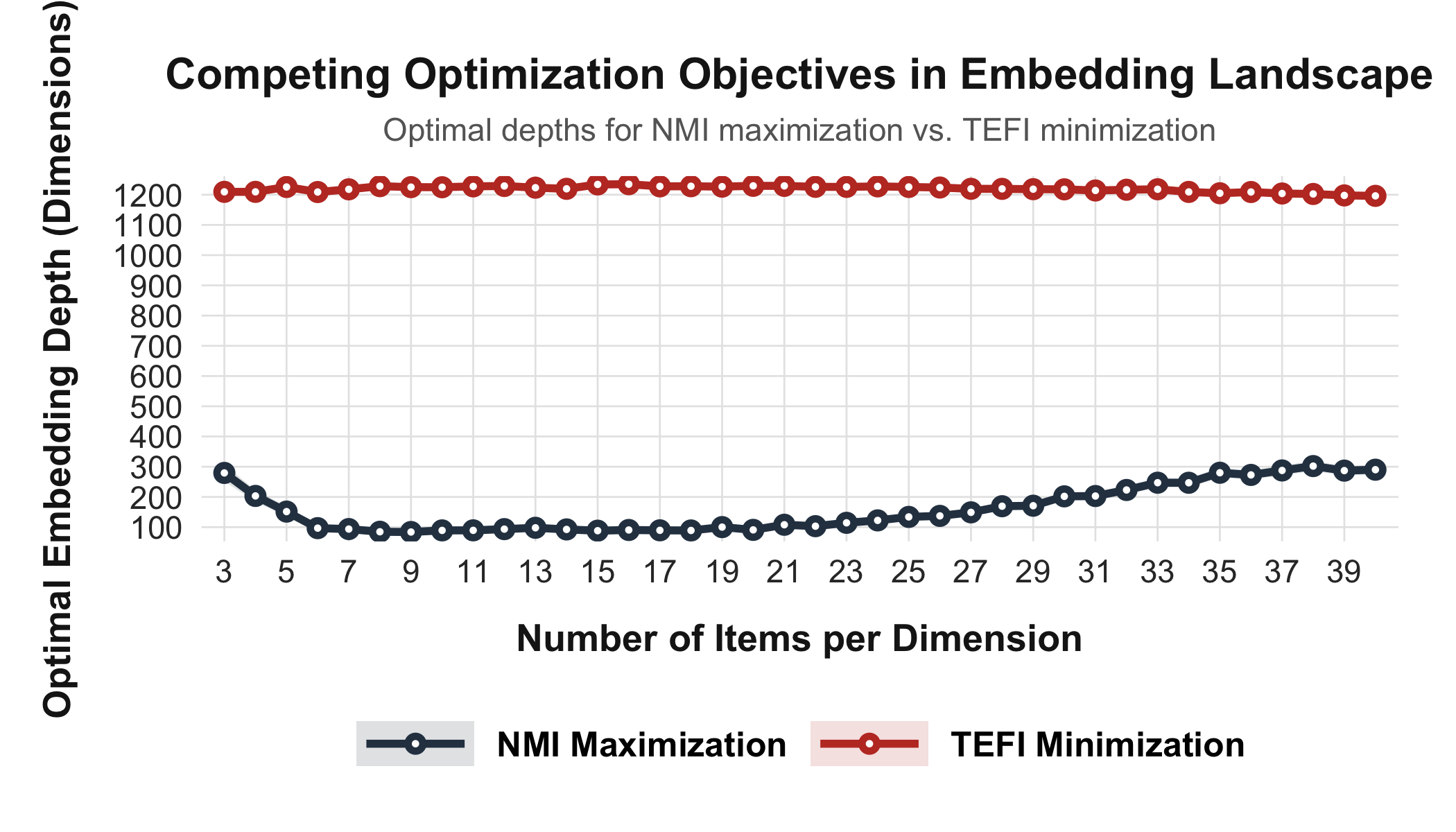}
\caption{Comparison of mean embedding dimension depth (y-axis) per number of items (x-axis) optimized via total entropy fit index (red circles) and normalized mutual information (blue circles).}
  \label{fig:comparison_plot}
\end{figure}

Figure ~\ref{fig:comparison_plot} shows that the optimal regions for TEFI is located at different depths than the optimal regions for NMI, confirming that single-metric optimization is insufficient. For item pools with 20-30 items per dimension, TEFI is optimized at depths around 1200 embedding dimensions, whereas NMI is optimized at around 100 embedding dimensions for similar item counts. This offset shows why the weighted composite metric is necessary. Neither metric alone identifies the balanced solution that jointly optimizes structural accuracy and entropy-based organization.

\begin{figure}[!ht]
  \centering
    \includegraphics[width=\linewidth]{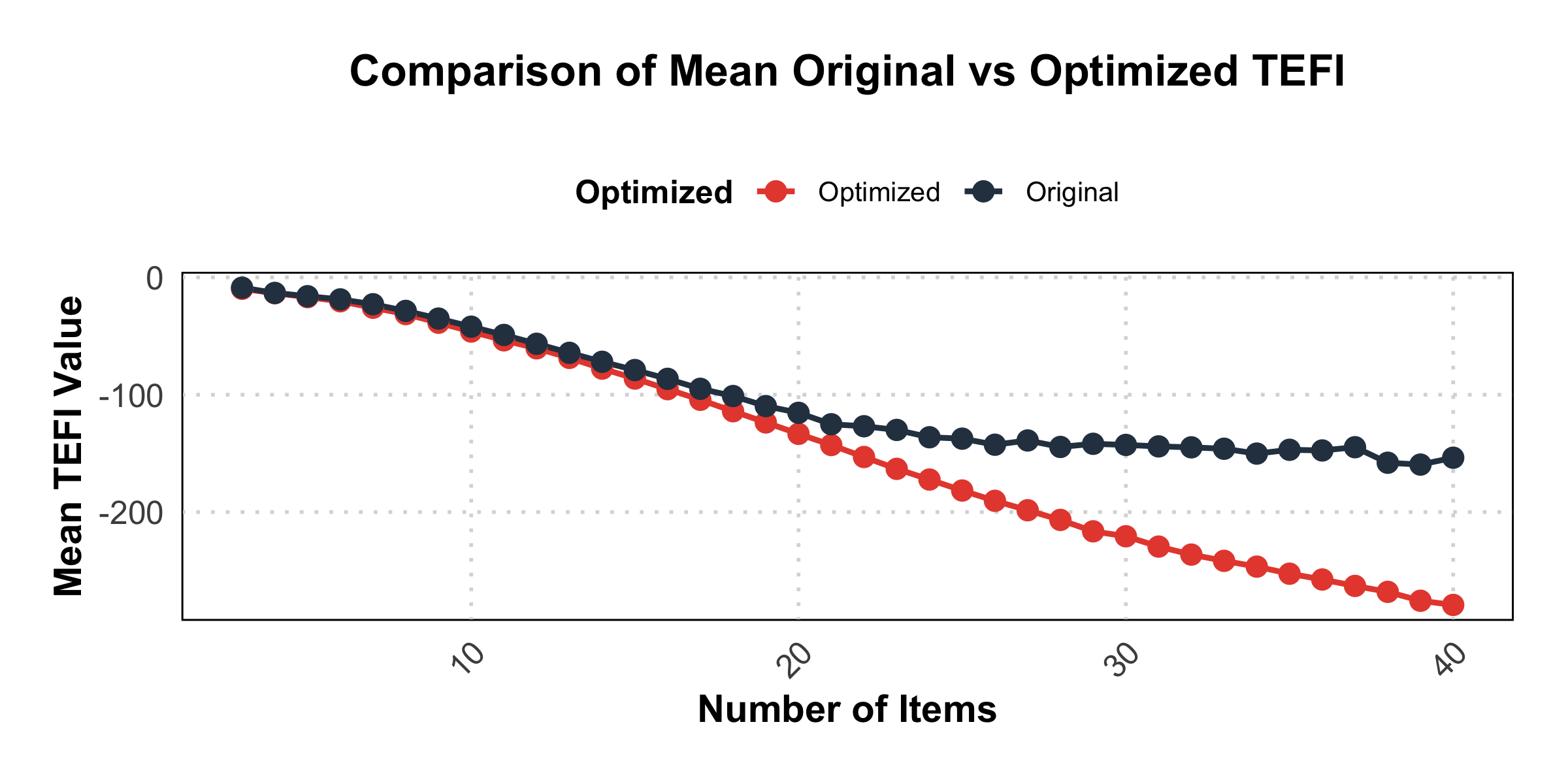}
\caption{Comparison of mean accuracy for the original (cross-sectional EGA of the embeddings) and optimized (Dynamic EGA of the embeddings searching along the number of embedding dimensions via the joint NMI-TEFI optimization) results.}
  \label{fig:accuracy_plot}
\end{figure}

Figure~\ref{fig:accuracy_plot} compares the mean structural accuracy for the original cross-sectional embedding analysis (as used in AI-GENIE) and the optimized embedding depths selected via the joint NMI–TEFI criterion. Across all item counts, the optimized solutions consistently achieve higher accuracy, with the largest gains observed for moderate to large item pools. These results demonstrate that searching the embedding landscape using a composite optimization strategy leads to systematic improvements in structural recovery relative to fixed, cross-sectional embedding representations.

\section{Discussion}

\subsection{Embedding Landscapes and Structural Non-Uniformity: Reconceptualizing Semantic Representations in Measurement}

The present findings demonstrate that LLM embedding spaces are structurally non-uniform and cannot be treated as homogeneous feature vectors without substantial loss of psychometric information. Different regions of the embedding dimension index support distinct trade-offs between structural accuracy and entropy-based organization, revealing that embeddings function as stratified semantic manifolds rather than uniform representational spaces. Shallow embedding coordinates capture coarse semantic distinctions that facilitate dimensional recovery, whereas deeper coordinates increasingly encode fine-grained semantic variation that may undermine coherent dimensional structure. No single embedding depth is universally optimal across item pool configurations, underscoring the need to treat embedding depth as a tunable hyperparameter rather than a fixed property of the model.

This finding directly addresses a critical gap in current generative psychometric practice. Existing frameworks such as AI-GENIE \citep{russell_2024aigenie, garrido2025estimating} and recent embedding-based measurement approaches \citep{wulff2025embeddinga} have established that embeddings can support dimensionality assessment and scale development prior to data collection, yet these applications implicitly assume that full embedding vectors should be analyzed wholesale. The landscape perspective challenges this assumption by demonstrating that optimal psychometric information is localized to specific coordinate ranges, and that indiscriminate use of all embedding dimensions introduces noise that degrades structural fidelity.

By conceptualizing embeddings as searchable landscapes traversed via Dynamic Exploratory Graph Analysis \citep{golino2020modeling}, the method reveals stable regions, transitional zones, and unstable regimes that are invisible under conventional cross-sectional Exploratory Graph Analysis \citep{golino2017ega1, golino2019investigating, christensen2020comparing}. The vector field results further show that trajectories in TEFI–NMI space are directionally structured, with attractor-like regions where both metrics change minimally and divergent regions where improvements in one metric are accompanied by degradation in the other. These dynamics clarify why single-metric optimization strategies are insufficient for principled embedding selection and establish multi-objective criteria as methodologically necessary rather than merely preferable.

\subsection{Item Pool Richness and Optimal Embedding Depth: Bridging Generation and Structure Recovery}

Item pool size (i.e., number of items per dimension/facet/construct) emerged as a critical determinant of where optimal structure resides in the embedding landscape. Sparse item pools require broader semantic regions to differentiate limited number of items, leading to optimal solutions at greater embedding depths. As item pool richness increases (more items per dimension/facet/construct), optimal embedding depth shifts toward shallower or intermediate regions, reflecting more efficient representation of construct structure. However, this trend saturates and, in some cases, reverses when item pools become very large, suggesting diminishing returns or redundancy effects that interfere with clustering accuracy.

These findings have direct implications for generative workflows like AI-GENIE \citep{russell_2024aigenie}, which use LLMs to generate candidate items and embeddings to estimate dimensionality before empirical validation. The results indicate that moderate item pool sizes (approximately 10–20 items per dimension) yield the most reliable structural recovery when paired with appropriately selected embedding depths. Very small pools lack sufficient semantic coverage to stabilize dimensional boundaries in embedding space, whereas very large pools may overload the representation with overlapping or paraphrastic content that obscures dimensional organization.

This finding converges with recent evidence from \citet{garrido2025estimating}, who demonstrated that LLM-generated item pools inherently contain semantic redundancy and that network-based filtering procedures (e.g., Unique Variable Analysis, bootstrap EGA) are essential for removing unstable or redundant/duplicative items. The landscape optimization approach complements these filtering methods by identifying the embedding depth ranges where filtered item pools exhibit maximal structural coherence. Rather than replacing content-based refinement procedures, landscape optimization operates synergistically with them. Filtering removes problematic items, while depth optimization extracts the most informative semantic signal from the refined pool.

The interaction between item pool size and optimal depth also provides insight into the capacity constraints of embedding spaces. The saturation effect observed at 30–40 items per dimension suggests that embeddings have finite representational bandwidth for differentiating construct-relevant semantic variation. Beyond this threshold, additional items do not enrich the semantic space but instead introduce redundancy that blurs dimensional boundaries. This constraint has practical importance for scale development workflows, indicating that targeting pools of 15–25 items per dimension balances semantic coverage against representational efficiency.

\subsection{Competing Optimization Objectives and the Necessity of Composite Criteria}

A central contribution of this work is the explicit demonstration that Normalized Mutual Information and the Total Entropy Fit Index \citep{golino2019entropy, golino2024gentefi} encode fundamentally different optimization objectives that often compete across the embedding landscape. NMI prioritizes correspondence with the generating dimensional structure, measuring clustering accuracy in terms of mutual information between estimated and true partitions. TEFI quantifies entropy-based organization using quantum information-theoretic principles \citep{vonneumann1927, quantuminfo}, penalizing disordered or overly complex dimensional solutions. Across the embedding landscape, these metrics often peak at markedly different depths, producing competing optima that cannot be jointly satisfied through single-metric optimization.

Optimizing either metric in isolation leads to problematic solutions with distinct failure modes. Shallow embeddings selected by maximizing NMI alone achieve acceptable clustering correspondence but poor entropy-based organization, yielding dimensionally accurate but informationally disordered structures. Conversely, deep embeddings selected by minimizing TEFI alone achieve strong entropy minimization and within-dimension coherence but at the cost of degraded structural accuracy, producing highly organized but dimensionally inaccurate partitions. Neither trajectory identifies the balanced intermediate solution that psychometric applications require.

The weighted composite criterion resolves this multi-objective optimization problem by identifying regions of the embedding space that jointly balance accuracy and organization. Importantly, the composite does not collapse the distinction between metrics but operationalizes their trade-off in a transparent and adjustable manner. The 70/30 weighting prioritizes structural accuracy while maintaining entropy-based coherence as a necessary constraint, reflecting the view that recovering the correct dimensional organization is the primary goal while ensuring that recovered dimensions are interpretable and informationally stable.

This approach aligns with broader principles in modern psychometric modeling, where model selection must balance multiple objectives including structural fidelity, parsimony, and interpretability. The TEFI itself was developed to address limitations of purely data-driven community detection algorithms that may produce statistically optimal but substantively incoherent dimensional solutions \citep{golino2019entropy}. By incorporating TEFI alongside structural accuracy metrics, the landscape optimization framework operationalizes this balance explicitly and systematically across embedding depth ranges.

\subsection{Embedding-Based Dimensionality Assessment: Advancing Beyond PCA Approximations}

These results extend recent methodological evidence that analytical choices in embedding-based dimensionality assessment have foundational rather than merely preliminary consequences. \citet{garrido2025estimating} demonstrated that principal component analysis applied to item embeddings systematically overestimates dimensionality, recovering inflated factor structures that misrepresent the generating dimensional organization. In contrast, network-based Exploratory Graph Analysis recovers generating structure with substantially greater accuracy, achieving near-perfect dimensional recovery in conditions where PCA produces solutions with twice the correct number of dimensions.

The present study advances this evidence in three critical ways. First, it reveals \textit{why} full-vector PCA approaches fail. They treat all embedding coordinates as equally informative and apply variance decomposition uniformly across the entire semantic space. This assumption is empirically false. Optimal structural information is concentrated in specific depth ranges, with early coordinates capturing coarse semantic organization and intermediate ranges providing optimal signal-to-noise ratios. PCA conflates signal-rich intermediate coordinates with noise-laden deep coordinates, producing variance patterns that suggest excessive dimensionality. Network-based approaches using partial correlations and community detection are more robust to this heterogeneity, but even standard cross-sectional EGA underperforms relative to optimized DynEGA solutions that explicitly locate and extract information from optimal depth ranges, as shown in Figure~\ref{fig:accuracy_plot}.

Second, the landscape approach demonstrates that even when using superior EGA-based network methods, \textit{how} embeddings are analyzed matters as much as \textit{which} method is used. Cross-sectional EGA applied to full embedding vectors, the current standard in generative psychometrics \citep{russell_2024aigenie, garrido2025estimating}, achieves higher accuracy than PCA but remains suboptimal relative to depth-optimized DynEGA. Systematic landscape optimization yields consistent improvements in both structural recovery (NMI) and entropy-based fit (TEFI) across all item pool sizes, with largest gains observed for pools of 10–20 items per dimension.

Third, the competing trajectories of TEFI and NMI formalize a conceptual distinction between two aspects of dimensional structure that existing methods typically conflate. The first is \textit{boundary accuracy} (how well items cluster according to their generating dimensions) and the second is \textit{internal coherence} (how informationally organized each dimension is). NMI captures the former, TEFI captures the latter, and their divergence demonstrates that achieving accurate boundaries does not guarantee coherent dimensions, nor does maximizing coherence ensure accurate boundaries. This distinction is psychometrically consequential because dimensional solutions may correctly identify the number and composition of dimensions (satisfying boundary accuracy) while producing dimensions that are internally heterogeneous, unstable, or difficult to interpret (failing internal coherence). Conversely, solutions may exhibit strong within-dimension organization while misallocating items across dimensions or recovering incorrect dimensionality. The composite optimization criterion explicitly balances these competing desiderata in ways that single-metric approaches cannot.

\subsection{Interpretability and the Architecture of Semantic Representation}

The embedding landscape framework provides a bridge between psychometric measurement and interpretability research in natural language processing. By mapping how structural metrics evolve across embedding depth, DynEGA offers insight into how representational capacity is allocated across the embedding space and how semantic organization emerges from transformer architectures.

The periodic oscillations observed in TEFI trajectories may reflect deeper architectural properties of transformer models, specifically how information propagates through attention layers, residual connections, and layer normalization operations. Transformer representations are known to organize semantic information hierarchically across layers, with early layers capturing syntactic and surface-level features and later layers encoding abstract semantic relationships \citep{OpenAIembeddings}. The embedding dimension index may partially preserve this layered organization, such that traversing embedding coordinates parallels traversing representational depth in the original model. Understanding these connections could inform prompt engineering, embedding postprocessing, and model fine-tuning strategies that enhance psychometric fidelity for specific measurement applications.

\subsection{Semantic Structure as Substantive Validity Evidence}

More broadly, the results reinforce an emerging reconceptualization in computational psychometrics, i.e., semantic structure encoded in embeddings should be treated as substantive validity evidence rather than merely a convenient proxy for later empirical covariation \citep{wulff2025embeddinga}. Recent work demonstrates that embedding-based item similarities can anticipate empirical item covariation patterns \citep{hommel2025language}, support cross-instrument prediction \citep{ravenda2025rethinking}, and constrain observed dimensional organization through wording and content decisions embedded in item generation. These findings suggest that dimensional structure recovered from embeddings is not an approximation awaiting empirical confirmation but a foundational component of construct definition that shapes what can be observed empirically.

This perspective has significant implications for construct validity. Embeddings alow taxonomic evaluation across multiple instruments and broad construct domains simultaneously, a scope of comparison that empirical validation cannot achieve because administering all candidate items together would be infeasible. Embedding-based network methods thus provide tools for assessing content validity and conceptual clarity by identifying dimensions that are semantically indistinguishable despite distinct construct labels, detecting drift of item sets toward adjacent constructs over instrument revisions, and revealing misalignment between theoretical construct definitions and actual item content \citep{wulff2025embeddinga}. These capabilities move beyond traditional content validity judgments based on expert ratings toward systematic, scalable, and reproducible evaluation of semantic structure.

By localizing where in the embedding space dimensional structure is most coherent, the landscape approach provides actionable guidance for item generation, selection, and refinement within generative frameworks like AI-GENIE \citep{russell_2024aigenie}. Rather than treating all embedding coordinates as equally informative or relying on default full-vector analyses, researchers can systematically identify optimal depth ranges for specific item pools and construct domains. This optimization is computationally straightforward. DynEGA can traverse embedding coordinates systematically and evaluate composite fit criteria at each depth, and yields substantial gains in both structural fidelity and entropy-based organization.

The method also addresses persistent taxonomic ambiguities in psychological measurement. Constructs that appear conceptually distinct may exhibit indistinguishable embedding-based dimensional structures, suggesting semantic overlap that empirical covariation alone cannot resolve. Conversely, constructs labeled identically across instruments may exhibit divergent embedding structures, indicating substantive differences in how construct domains are operationalized through item content. These insights have implications for measurement equivalence, cross-instrument harmonization, and theoretical integration across research traditions that use different measurement tools for ostensibly identical constructs.

\subsection{Limitations and Future Directions}

Several limitations warrant consideration. First, the simulations focused on a single construct domain (grandiose narcissism) with known five-dimensional structure. Whether patterns generalize to hierarchical models or bifactor structures remains an empirical question. The periodic oscillations in TEFI and the specific depth ranges where NMI peaks may be construct-specific or may reflect universal properties of how embedding models organize semantic content. Systematic replication across diverse constructs, such as personality traits, psychopathology symptoms, attitude domains, cognitive abilities, is necessary to distinguish domain-general principles from construct-specific patterns.

Second, the 70/30 weighting for the composite criterion was chosen to prioritize structural accuracy based on the assumption that recovering correct dimensional organization is the primary measurement goal. However, alternative weights may be appropriate when within-dimension coherence is valued over between-dimension differentiation, when theoretical interest focuses on entropy minimization rather than dimensional recovery, or when measurement contexts require maximally stable and interpretable dimensions even at the cost of some clustering accuracy. Formal Pareto frontier analysis mapping the complete trade-off space in (TEFI, NMI) coordinates would provide more systematic guidance for weight selection under varying measurement priorities.

Third, only one embedding models was examined, and the broader landscape of available architectures, including Cohere, Google's Universal Sentence Encoder, and domain-specialized sentence transformers, may exhibit different representational strategies with different optimal depth profiles. Extending the landscape framework to additional models would clarify whether the observed patterns (adaptive capacity redistribution vs. constrained compression) represent fundamental architectural trade-offs or model-specific design choices that could be optimized through training procedures.

Fourth, the study focused on pre-empirical dimensionality assessment using generated items without empirical response data. Future work should validate that embedding-based structures optimized via DynEGA correspond to empirically estimated networks from actual response data \citep{golino2017ega1}. Such alignment studies would establish whether landscape optimization genuinely improves construct measurement or merely optimizes embedding-internal criteria that may not translate to empirical validity. Integration with recent work showing that embeddings can anticipate empirical covariation \citep{hommel2025language, ravenda2025rethinking} suggests that optimized embedding structures should exhibit stronger empirical correspondence, but direct validation remains necessary.

Fifth, the connection between embedding depth and transformer architecture, specifically how the embedding dimension index relates to layer-wise representations, attention patterns, and information flow through the model—remains theoretically underspecified. Future work combining DynEGA landscape analysis with layer-wise activation studies and attention weight visualizations could reveal how dimensional structure emerges at different representational depths and inform model development targeted specifically at psychometric applications.

Finally, the study employed Triangulated Maximally Filtered Graphs \citep{massara2016network} and the Walktrap algorithm \citep{pons2006walktrap} for network estimation and community detection. Alternative network construction or estimation methods (e.g., graphical LASSO) and community detection algorithms (e.g., Louvain) may produce different landscape patterns and optimal depth ranges. Systematic comparison of methodological variants would establish the robustness of landscape optimization principles across analytical choices.

\subsection{Future Research Directions}

Beyond addressing limitations, several promising directions emerge from this work. \textbf{Pareto optimization studies} should map complete trade-off frontiers in (TEFI, NMI) space to provide principled guidance for weight selection under different measurement priorities, potentially revealing that certain item pool configurations admit Pareto-dominant solutions where both metrics can be jointly improved without trade-offs. \textbf{Cross-construct validation studies} should test landscape patterns across diverse domains to establish boundary conditions and identify construct-specific optimal depth profiles. \textbf{Empirical-embedding alignment studies} should compare DynEGA-optimized embedding networks with networks estimated from empirical response data to validate that semantic structure optimization translates to improved psychometric performance. \textbf{Layer-wise activation analyses} should connect embedding landscape patterns to internal transformer representations to establish mechanistic links between model architecture and dimensional structure emergence. \textbf{Longitudinal construct drift studies} should apply landscape optimization to historical item pools to detect how construct operationalizations evolve over time and instrument revisions. \textbf{Cross-lingual measurement equivalence studies} should extend the framework to multilingual embeddings to assess whether dimensional structures remain stable across languages and cultural contexts.

\section{Conclusion}

By treating LLM embeddings as dynamic, searchable landscapes rather than static feature vectors, Dynamic Exploratory Graph Analysis provides a principled framework for optimizing embedding usage in generative psychometrics. Large-scale Monte Carlo simulations demonstrate that embedding depth is a consequential modeling choice with systematic effects on both structural recovery and entropy-based organization, and that landscape optimization yields consistent improvements over conventional cross-sectional approaches. The method reveals that embedding spaces are non-uniform, with optimal psychometric information localized to specific coordinate ranges that vary as a function of item pool richness.

These findings challenge current practice in embedding-based dimensionality assessment, which treats full embedding vectors as default representations and applies PCA or cross-sectional EGA without systematic depth optimization. The evidence indicates that such approaches are demonstrably suboptimal, conflating signal-rich intermediate coordinates with noise-laden deep coordinates and failing to account for the structured organization of semantic information across embedding dimensions. The landscape optimization framework positions embedding analysis not as a preliminary heuristic awaiting empirical validation, but as a foundational methodological component of modern measurement theory that provides substantive validity evidence at scales that empirical data collection cannot achieve.

More broadly, this work advances the integration of natural language processing and psychometric theory, demonstrating that computational tools like embeddings and transformers are not merely convenient approximations but substantive measurement technologies that shape what constructs mean and how they can be operationalized. As generative psychometric workflows like AI-GENIE \citep{russell_2024aigenie} become increasingly central to scale development, the principles established here—systematic depth optimization, multi-objective criteria, model-aware analysis strategies—provide methodological foundations for principled, reproducible, and theoretically grounded measurement in the computational era.

\paragraph{Funding Statement}
This research received funding from the Jefferson Trust.

\paragraph{Competing Interests}
The author declares no competing interests.

\printbibliography

\end{document}